# Locally Deployable Small Language Models for Emergency Department Decision Support: A Systematic Benchmark of Fine-Tuning Strategies

**Qingfeng Zhang, M.S.[1], Yuanxiong Guo, Ph.D.[1], Yanmin Gong, Ph.D.[2]**
**[1]The University of Texas at San Antonio, San Antonio, TX; [2]Texas A&M University, Houston, TX**

**Abstract**
*Deploying large language models (LLMs) for decision support in emergency departments (EDs) faces two major challenges: privacy risks of transmitting patient data to closed-source commercial LLMs and the lack of systematic evaluation of fine-tuning strategies for locally deployable open-source small language models (SLMs). We benchmarked eight open-source SLMs using zero-shot prompting, prefix tuning, Low-Rank Adaptation (LoRA), and full fine-tuning on three ED tasks: triage level prediction, specialist referral recommendation, and diagnosis prediction. Using 2,083 MIMIC-IV-ED cases and Claude Haiku 4.5 and Claude Sonnet 4.5 as baselines, we found that LoRA fine-tuned open-source SLMs outperform commercial baselines on triage level prediction and specialist referral recommendation, while diagnosis prediction remains challenging for open-source SLMs. Confusion matrix analysis further shows that fine-tuned open-source SLMs can detect highest-severity patients missed by the commercial baselines. These results demonstrate that locally deployable SLMs can achieve clinically competitive performance for ED decision support.*

## Introduction

Accurate and timely clinical decision-making is among the most consequential responsibilities in medicine. In emergency department (ED) settings, clinicians must rapidly integrate patient symptoms, vital signs, and medical history to determine triage priority, recommend appropriate specialist referral, and formulate diagnoses, frequently under extreme time pressure and with incomplete information. Errors in any of these decisions may delay necessary care, lead to inappropriate consultations, or misuse limited clinical resources. For example, a systematic review of field triage studies reported under-triage rates of 14–34% and over-triage rates of 12–31%[1], both of which adversely impact resource allocation and patient outcomes. As healthcare systems face growing patient demand with a constrained clinical workforce, decision support tools that help clinicians make these decisions faster and more reliably are increasingly needed.

Recent studies have demonstrated the promise of large language models (LLMs) in clinical decision support[2,3,4]. Beyond strong performance on structured medical benchmarks such as medical licensing exams[5,6], several studies have now evaluated LLMs on real-world, open-ended clinical tasks. Commercial LLMs have shown clinically meaningful accuracy in ED triage prediction, specialist referral recommendation, and diagnosis generation using real patient cases[7], provide reliable clinical recommendations and acuity assessments across large ED cohorts[8,9], and match or exceed physician-level performance in clinical text summarization[10]. A scoping review has further surveyed the growing role of LLMs across multiple aspects of emergency medicine[11]. These findings suggest that LLMs hold potential to support clinical decision-making in the ED.

Despite these promising results, two challenges limit the deployment of LLMs in hospitals. First, commercial models such as GPT, Claude, and Gemini are commonly accessed through cloud-based application programming interface (API) services, which require patient data to be transmitted outside the hospital. This raises privacy, governance, and data sovereignty concerns under regulations such as Health Insurance Portability and Accountability Act of 1996 (HIPAA) and General Data Protection Regulation (GDPR)[12]. Beyond regulatory compliance, cloud-based API services may also introduce recurring costs, vendor dependence, and network latency[13]. Second, few studies have systematically compared adaptation methods[14,15] for open-source small language models (SLMs) across multiple ED decision-support tasks. In this study, we use SLMs to refer to open-source models with 1–8 billion parameters, a scale that is more suitable for local deployment under hospital resource constraints than larger commercial LLMs. These methods include zero-shot prompting, prefix tuning, Low-Rank Adaptation (LoRA)[16], and full fine-tuning. It therefore remains unclear which adaptation method offers the best balance between performance and computational cost for local deployment. These challenges motivate a systematic evaluation of locally deployable SLMs and practical adaptation strategies under realistic hospital constraints.

In this study, we systematically compare prompting and fine-tuning strategies for open-source SLMs on three ED decision-support tasks: triage level prediction in the form of the Emergency Severity Index (ESI)[17], specialist referral recommendation, and diagnosis prediction. Using 2,083 cases derived from the Medical Information Mart for Intensive Care (MIMIC-IV) database[18], we evaluate eight SLMs spanning 1B to 8B parameters. Each model is assessed using zero-shot prompting and three different fine-tuning methods: prefix tuning, LoRA, and full fine-tuning. In addition to overall accuracy, we examine triage confusion matrices to characterize under-triage and over-triage across ESI levels. Our results show that task-specific fine-tuning substantially improves SLM performance. Several adapted SLMs achieve higher accuracy than the commercial zero-shot baselines on triage and specialist referral, while diagnosis prediction remains more challenging.

## Related Work

**LLMs in ED Decision Support** Recent studies have evaluated commercial LLMs for several ED decision support tasks, but their reliance on cloud-based API services may limit their deployment in hospital settings. Gaber et al. have tested Claude-family models and a retrieval-augmented generation (RAG) workflow on 2,000 MIMIC-IV cases across triage, referral, and diagnosis prediction[7]. Williams et al. have demonstrated GPT-4-level accuracy in pairwise ESI acuity comparisons and clinical recommendation generation across large ED cohorts[8,9], and Van Veen et al. have shown that adapted LLMs can surpass physician performance in clinical text summarization[10]. However, transmitting patient data to external cloud servers raises concerns about privacy, data governance, and compliance with HIPAA and GDPR requirements[12]. Dennstädt et al. have discussed that locally hosted LLMs provide superior institutional data control[13], and Riedemann et al. have further argued that the future of clinical AI must rest on transparent open-source models that clinicians can audit and hold accountable[19]. However, few studies have examined whether open-source models adapted for ED tasks can perform as well as commercial LLMs.

**Adaptation of SLMs in Healthcare** SLMs require less memory and computation than larger models, making local deployment more practical. However, their zero-shot performance on clinical tasks often falls behind that of larger models. Fine-tuning can improve task-specific performance, but different methods vary in the number of parameters they update. Prefix tuning adds trainable vectors to each transformer layer while keeping the model weights fixed[20], whereas LoRA introduces low-rank updates to selected model layers[16]. Ding et al. have shown that LoRA consistently achieves the strongest performance-efficiency trade-off among fine-tuning approaches[21]. Full fine-tuning, in contrast, updates all model parameters. Previous studies have applied fine-tuning to clinical tasks. Gema et al. have shown that a fine-tuning framework improved International Classification of Diseases (ICD) diagnosis classification over BERT-based clinical models[22]. Zhang et al. have demonstrated that LoRA fine-tuning of Llama 2 reduces the gap with GPT-4 for medical evidence summarization[23], and Gao et al. have shown that LoRA fine-tuned 4B–32B open-source models reach expert level perioperative complication detection with complete data sovereignty[24]. However, to our knowledge, no prior studies have systematically compared multiple fine-tuning strategies across several SLM families using the same set of ED decision-support tasks.

## Methods

**Task Overview** Following the benchmark established by Gaber et al.[7], we evaluate open-source SLMs on three ED decision support tasks based on patient symptoms, demographics, and vital signs: *triage level prediction* to assign an ESI score from 1 to 5, *specialist referral recommendation* to predict the top-three appropriate medical specialties, and *diagnosis prediction* to generate the top-three most likely primary diagnoses. To address the need for privacy-preserving clinical decision support, we systematically benchmark eight open-source SLMs under zero-shot evaluation and three different fine-tuning methods: prefix tuning, LoRA, and full fine-tuning. For each model and fine-tuning strategy combination, we compared task accuracy against zero-shot baselines established by Claude Haiku 4.5 and Claude Sonnet 4.5. This comparison assessed whether locally fine-tuned SLMs could match the performance of commercial models. Additionally, we analyze triage confusion matrices to examine under-triage and over-triage patterns across ESI levels, providing a more comprehensive assessment of model performance and clinical safety.

**Dataset** Following the preprocessing framework of Gaber et al.[7], we constructed the study cohort from the MIMIC-IV-ED and MIMIC-IV-Note datasets[25,26]. The final cohort included 2,083 ED cases. The distribution of ESI levels in the final cohort is shown in Table 1. Specifically, we merged the MIMIC-IV-ED triage table, edstays table, and patient table with discharge notes from MIMIC-IV-Note. Triage labels and initial vital signs were extracted from the triage file, demographic variables including sex and race were obtained from the edstays file, and age was taken from the patient table. Discharge notes were linked to ED encounters using stay_id. During preprocessing, duplicate subject entries were removed, cases with missing triage information were excluded, only encounters with seq_num = 1 were retained to ensure unique ED stays, and patients marked as deceased were excluded. For unstructured notes, we retained only cases with a documented history of present illness (HPI) and extracted the HPI paragraph while discarding the remainder of the discharge note. We kept HPIs between 50 and 2000 characters, removed nonspecific entries that did not meaningfully describe presenting symptoms, and extracted primary diagnoses from the notes. This preprocessing is intended to preserve clinically relevant information from initial visits while reducing note noise.

Table 1: Distribution of ESI levels

| Triage Level (ESI) | Patient Count (N) | Percentage (%) |
|---|---|---|
| 1 | 171 | 8.21% |
| 2 | 962 | 46.18% |
| 3 | 938 | 45.03% |
| 4 | 12 | 0.58% |
| 5 | 0 | 0.00% |

Each final case contains a patient's HPI, demographic information, and initial triage vital signs. Ground-truth triage labels correspond to the documented ESI level, and diagnosis labels correspond to the primary discharge diagnoses extracted from the notes. Because MIMIC-IV does not include specialist referral labels, we generated reference specialty labels from the primary diagnoses using Claude Sonnet 4.5, following Gaber et al.[7]. We split the 2,083 cases into training, validation, and test sets in a 60/20/20 ratio. All zero-shot and fine-tuned models were evaluated on the same test set to ensure a consistent comparison. The datasets are not publicly available in their processed form, which reduces the risk that any model has been exposed to these cases during pretraining.

You are a nurse with emergency and triage experience. Using the patient's history of present illness and his information, determine the triage level based on the Emergency Severity Index (ESI), ranging from ESI level 1 (highest acuity) to ESI level 5 (lowest acuity):
1: Assign if the patient requires immediate lifesaving intervention.
2: Assign if the patient is in a high-risk situation (e.g., confused, lethargic, disoriented, or experiencing severe pain/distress)
3: Assign if the patient requires two or more diagnostic or therapeutic interventions and their vital signs are within acceptable limits for non-urgent care.
4: Assign if the patient requires one diagnostic or therapeutic intervention (e.g., lab test, imaging, or EKG).
5: Assign if the patient does not require any diagnostic or therapeutic interventions beyond a physical exam (e.g., no labs, imaging, or wound care).

History of present illness: {HPI}, patient info: {patient_info} and initial vitals: {initial vitals}
Respond with the level in an <acuity> tag.

(a) Triage level prediction prompt

You are an experienced healthcare professional with expertise in determining the medical specialty and diagnosis based on a patient's history of present illness and personal information. Review the data and identify the three most likely, distinct specialties to manage the condition, followed by the three most likely diagnoses. List specialties first, in order of likelihood, then diagnoses.

History of present illness: {hpi}, personal information: {patient_info} and initial vitals: {initial vitals}
Respond with the specialties in <specialty> tags and the diagnoses in <diagnosis> tags.

(b) Joint referral and diagnosis prediction prompt

You are an experienced healthcare professional with expertise in medical and clinical domains. I will provide a list of real diagnoses for a patient and 3 predicted diagnoses. For each predicted diagnosis, determine if it has the same meaning as one of the real diagnoses or if the prediction falls under a broader category of one of the real diagnoses (e.g., a specific condition falling under a general diagnosis category). If it matches, return 'True'; otherwise, return 'False'. Return only 'True' or 'False' for each predicted diagnosis within <evaluation> tags and nothing else.

Real Diagnoses: {real_diag}, predicted diagnosis 1: {diag1}, predicted diagnosis 2: {diag2}, and predicted diagnosis 3: {diag3}.

(c) LLM-as-a-judge prompt for diagnosis evaluation

**Figure 1.** Prompt templates for three prediction tasks and diagnosis evaluation

**Prompt Design** As shown in Figure 1(a) and 1(b), we designed a prompt template for triage level prediction and a joint prompt template for specialist referral and diagnosis prediction. All prompts provided the model with the patient's HPI, demographic information, and initial vital signs, and required structured outputs for reliable parsing. Each prediction prompt began with a role instruction framing the model as an experienced healthcare professional or triage expert, followed by the task description, patient data, and output format. For triage prediction, the model was instructed to return a single ESI level from 1 to 5 within <acuity> tags. For specialist referral and diagnosis prediction, the model was asked to generate the three most likely specialties and diagnoses within <specialty> and <diagnosis> tags. Because referral recommendation and diagnosis prediction are closely related in early ED decision making, we generated them jointly within a single prompt. We also used an LLM-as-a-judge prompt template for diagnosis evaluation, as shown in Figure 1(c). This prompt template compared each predicted diagnosis with the reference diagnosis set and determined whether it was clinically matching or semantically equivalent, returning binary outputs within <evaluation> tags. We tested several prompt templates on a development subset and selected the versions that produced the most consistent and reliably parsed outputs. The development cases were excluded from the final evaluation.

**Models and Adaptation Methods** We evaluated eight open-source SLMs spanning 1B to 8B parameters across three model families. From Meta's Llama family, we evaluated Llama-3.2-1B-Instruct, Llama-3.2-3B-Instruct, and Llama-3.1-8B-Instruct, chosen for their strong foundational knowledge base and logical reasoning. From Alibaba's Qwen family, we evaluated Qwen2.5-1.5B-Instruct, Qwen2.5-3B-Instruct, Qwen2.5-7B-Instruct, and Qwen3-4B-Instruct-2507, which demonstrate superior accuracy and strict adherence to structured output instructions among models of their size. From Mistral AI, we evaluated Mistral-7B-Instruct-v0.2, included for its proven balance between low computational overhead and complex text comprehension. Hereafter, we refer to Qwen3-4B-Instruct-2507 and Mistral-7B-Instruct-v0.2 as Qwen3-4B-Instruct and Mistral-7B-Instruct, respectively. These model families were selected to represent the most widely adopted open-source SLM architectures across a range of parameter scales relevant to resource-constrained hospital deployment. We used instruction-tuned variants to improve compliance with the structured output formats required for the three decision-support tasks.

To contextualize the performance of locally deployable open-source SLMs against strong proprietary systems, we additionally included Claude Haiku 4.5 and Claude Sonnet 4.5 as commercial baselines. According to Anthropic documentation available at the time of manuscript preparation, Claude Haiku 4.5 is their fastest and most cost-efficient model, designed for latency-sensitive and scaled applications while still providing strong reasoning and agentic capabilities. Claude Sonnet 4.5 is a substantially more capable model, with stronger performance on complex reasoning, coding, computer use, and agent workflows. In our benchmark, Haiku 4.5 serves as an efficient proprietary baseline representing lightweight, responsive deployment, whereas Sonnet 4.5 serves as a stronger reference for high-performance clinical decision support. Both models were evaluated in the zero-shot setting only.

Each SLM was evaluated using zero-shot prompting, prefix tuning, LoRA, and full fine-tuning. Zero-shot prompting used the original instruction-tuned model without updating its parameters. For prefix tuning, trainable virtual tokens were added while the model weights remained fixed. LoRA applied trainable low-rank matrices to selected attention projection layers, whereas full fine-tuning updated all model parameters. We used the same training, validation, and test sets for every model and adaptation method.

**Evaluation** We used the same evaluation framework of Gaber et al.[7]. For triage level prediction, we reported exact-match accuracy and range accuracy. Exact-match accuracy required the predicted ESI level to match the reference level. Range accuracy also accepted a prediction as correct if it equals the true level or is exactly one level more severe, with the exception that ESI-1 must be predicted exactly. For example, a reference ESI level of 3 was considered correct if the model predicted an ESI level of 2 or 3. It allows limited over-triage but does not accept under-triage.

For specialist referral recommendation, we report matched accuracy and at-least-one accuracy, following Gaber et al.[7]. For each case, matched accuracy was calculated for each patient by dividing the number of correctly predicted specialties by the length of the shorter list between the model's predictions and the ground truth. These individual patient scores are then averaged across the entire cohort to compute the overall matched accuracy. At-least-one accuracy considered a case correct if at least one of the three predicted specialties matched a reference specialty.

For diagnosis prediction, we followed the LLM-as-a-judge evaluation framework used by Gaber et al.[7], but replaced their commercial judge LLM with a locally deployed Llama-3.1-70B model. As a large open-source instruction-tuned model, Llama-3.1-70B provides strong reasoning ability and sufficient capacity to assess semantic and clinical correspondence between predicted and reference diagnoses, making it well suited for diagnosis evaluation in our benchmark. In this setup, Llama-3.1-70B evaluates each predicted diagnosis against the ground-truth list, returning True if the prediction matches or falls within a clinically equivalent broader category. We then calculated matched accuracy and at-least-one accuracy using the same method as for specialist referral. All judge inference was performed locally without transmitting patient data to external cloud services.

**Implementation Details** All experiments were conducted on a single server with four NVIDIA RTX A6000 GPUs. We used the AdamW optimizer, a batch size of 4, and bfloat16 precision. The learning rate was $1 \times 10^{-4}$ for prefix tuning and $2 \times 10^{-5}$ for LoRA and full fine-tuning. All adapted models were trained for 3 epochs. For LoRA, we used rank $r$ = 16, scaling factor $\alpha$ = 32, and dropout = 0.05. LoRA was applied to the attention projection layers $q_proj$, $k_proj$, $v_proj$, and $o_proj$. For prefix tuning, we used 20 virtual tokens with prefix projection enabled. For full fine-tuning, all model parameters were updated. All experiments were repeated over 5 random seeds, and results are reported as mean ± standard deviation. All models were evaluated under the same train/validation/test split for fair comparison. To maintain strict evaluation equity, all zero-shot baselines were also evaluated exclusively on this identical test set.

**Results**

**Triage Level Prediction** Tables 2 and 3 present triage prediction performance under exact-match and range accuracy, respectively, with Claude Haiku 4.5 and Claude Sonnet 4.5 serving as commercial zero-shot baselines. Among open-source SLMs, zero-shot performance varied substantially with model scale. For exact-match accuracy, Llama-3.1-8B-Instruct and Mistral-7B-Instruct achieved the strongest open-source zero-shot results at 53.59%, whereas smaller models such as Qwen2.5-1.5B-Instruct and Llama-3.2-1B-Instruct performed markedly worse at 27.51% and 29.43%. This pattern suggests that the larger models generally performed better on zero-shot triage prediction.

**Table 2:** Exact-match accuracy for triage level prediction across zero-shot prompting, prefix tuning, LoRA, and full fine-tuning. Results for trainable methods are reported as mean ± standard deviation over five seeds.

| Model | Zero-shot | Prefix tuning | LoRA | Full fine-tuning |
|---|---|---|---|---|
| Llama-3.2-1B-Instruct | 29.43 | 54.58 ± 4.42 | 60.94 ± 2.16 | 65.25 ± 1.88 |
| Qwen2.5-1.5B-Instruct | 27.51 | 48.57 ± 2.55 | 53.49 ± 3.32 | 64.32 ± 2.69 |
| Qwen2.5-3B-Instruct | 46.89 | 53.11 ± 2.87 | 65.50 ± 2.58 | 61.77 ± 3.49 |
| Llama-3.2-3B-Instruct | 41.63 | 51.77 ± 6.30 | 63.49 ± 2.43 | 62.73 ± 1.92 |
| Qwen3-4B-Instruct | 48.06 | 64.31 ± 1.81 | **65.98 ± 1.65** | 65.36 ± 2.27 |
| Mistral-7B-Instruct | 53.59 | 46.51 ± 0.97 | 65.74 ± 1.62 | 62.39 ± 2.05 |
| Qwen2.5-7B-Instruct | 51.91 | **60.14 ± 2.88** | 63.44 ± 1.95 | **65.57 ± 3.33** |
| Llama-3.1-8B-Instruct | 53.59 | 46.65 ± 1.76 | 62.58 ± 1.19 | 61.36 ± 1.61 |
| Claude Haiku 4.5 | 57.89 | / | / | / |
| Claude Sonnet 4.5 | **62.68** | / | / | / |

**Table 3:** Range accuracy for triage level prediction across zero-shot prompting, prefix tuning, LoRA, and full fine-tuning. Results for trainable methods are reported as mean ± standard deviation over five seeds.

| Model | Zero-shot | Prefix tuning | LoRA | Full fine-tuning |
|---|---|---|---|---|
| Llama-3.2-1B-Instruct | 42.82 | 74.84 ± 6.30 | 76.98 ± 6.52 | 79.13 ± 1.97 |
| Qwen2.5-1.5B-Instruct | 39.19 | 73.71 ± 5.23 | 70.10 ± 3.56 | 78.88 ± 1.87 |
| Qwen2.5-3B-Instruct | 46.03 | 75.79 ± 3.84 | 80.29 ± 2.14 | 77.66 ± 3.74 |
| Llama-3.2-3B-Instruct | 42.11 | 74.74 ± 7.02 | 78.23 ± 3.50 | 78.18 ± 2.96 |
| Qwen3-4B-Instruct | 50.33 | 78.71 ± 3.02 | **80.62 ± 2.64** | **80.38 ± 1.88** |
| Mistral-7B-Instruct | 57.89 | **79.38 ± 4.39** | 80.05 ± 2.61 | 77.92 ± 2.08 |
| Qwen2.5-7B-Instruct | 66.27 | 77.70 ± 2.70 | 79.23 ± 3.51 | 80.13 ± 1.88 |
| Llama-3.1-8B-Instruct | 71.53 | 75.98 ± 5.95 | 79.86 ± 1.86 | 77.28 ± 1.75 |
| Claude Haiku 4.5 | 76.79 | / | / | / |
| Claude Sonnet 4.5 | **82.54** | / | / | / |

After examining zero-shot performance, we evaluated triage performance using exact-match accuracy. Under this metric, adaptation substantially improved performance across all model families. Qwen3-4B-Instruct with LoRA achieved the highest observed accuracy at 65.98% ± 1.65%. This score exceeded Claude Sonnet 4.5, which achieved 62.68%, and Claude Haiku 4.5, which achieved 57.89%. Mistral-7B-Instruct with LoRA at 65.74% ± 1.62%, Qwen2.5-7B-Instruct with full fine-tuning at 65.57% ± 3.33% and Llama-3.2-1B-Instruct with full fine-tuning at 65.25% ± 1.88% also achieved higher observed accuracy than Claude Sonnet 4.5. These results show that, for exact ESI level triage prediction, fine-tuned open-source SLMs can match or even surpass strong commercial zero-shot baselines while remaining fully locally deployable. LoRA produced consistently strong results across several model families, although full fine-tuning performed best for some of the smaller models.

We next examined range accuracy. The results differed under range accuracy, which also accepted predictions that were one ESI level more urgent than the reference. Claude Sonnet 4.5 achieved the highest range accuracy at 82.54%, followed by Qwen3-4B-Instruct with LoRA at 80.62% ± 2.64% and full fine-tuning at 80.38% ± 1.88%. Mistral-7B-Instruct with LoRA reached 80.05% ± 2.61%, while Llama-3.1-8B-Instruct with LoRA reached 79.86% ± 1.86%. Thus, adapted SLMs achieved higher exact-match accuracy in several settings, but Claude Sonnet 4.5 retained the highest range accuracy. The smaller gap between the adapted SLMs and Claude under this metric indicates that many triage errors differed from the reference by only one ESI level. Because range accuracy accepts limited over-triage but not under-triage, it provides a complementary view of clinically conservative predictions.

To further characterize the clinical safety profile of triage predictions, Figure 2 compares the confusion matrices of Mistral-7B-Instruct with LoRA and Claude Sonnet 4.5 on the same test set. Neither model assigned any true ESI-1 case to ESI-4 or ESI-5, neither assigned a true ESI-4 case to ESI-1, and only 3 of 33 ESI-1 cases were misclassified by two levels to ESI-3. The lack of severe errors that skip multiple triage levels suggests that both the fine-tuned open-source model and the commercial baseline maintain a basic understanding of clinical urgency.

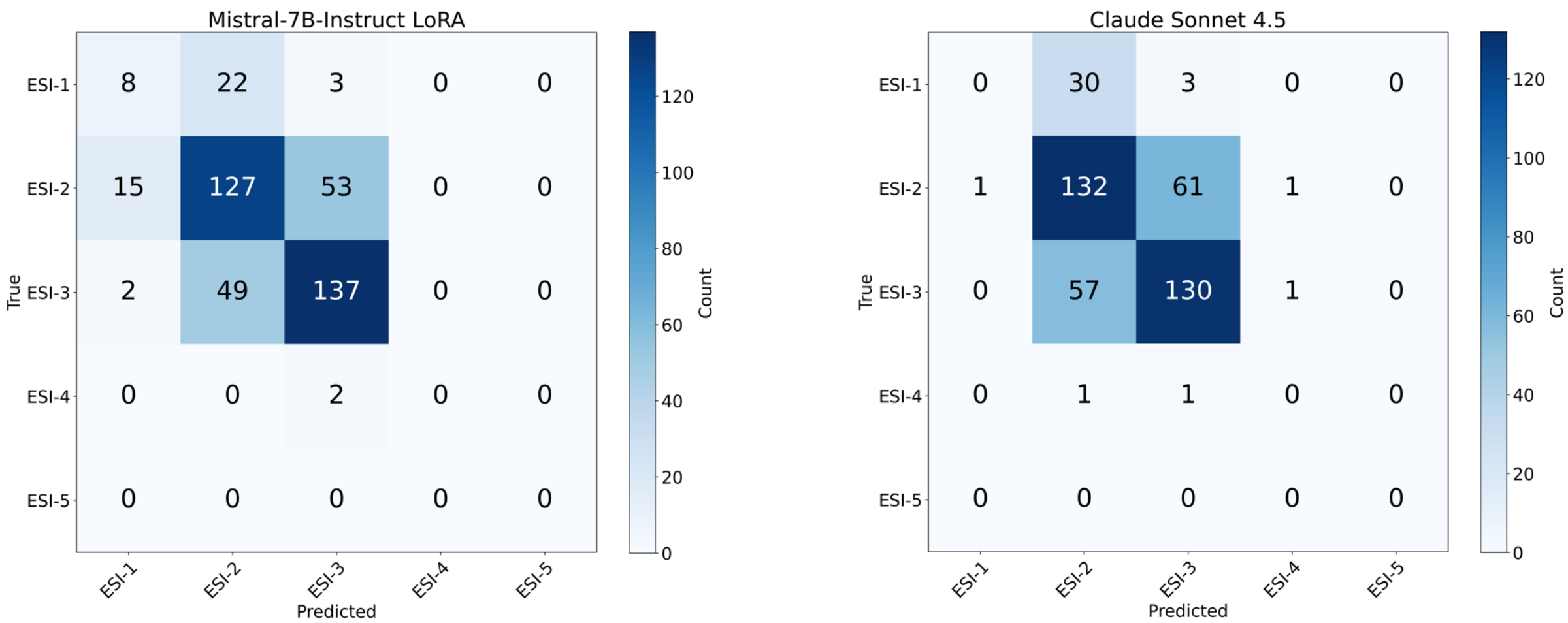

**Figure 2.** Triage level prediction confusion matrices for Mistral-7B-Instruct with LoRA and Claude Sonnet 4.5

The most clinically important difference lies in detection of the highest acuity patients. Mistral-7B-Instruct with LoRA correctly identified 8 of 33 true ESI-1 cases, corresponding to a sensitivity of 24.2%, while assigning most of the remaining cases to ESI-2. By contrast, Claude Sonnet 4.5 did not correctly identify any ESI-1 case and instead assigned 30 cases to ESI-2 and 3 cases to ESI-3. Although predictions of ESI-2 are still counted as acceptable under the range accuracy metric, complete failure to detect any ESI-1 case is clinically concerning because it misses the patients requiring the most immediate attention. This exploratory analysis suggests that aggregate accuracy alone may not fully describe triage performance and should be considered together with performance at individual ESI levels, and task-specific adaptation may improve the recognition of highest-acuity cases.

**Specialist Referral Recommendation** Tables 4 and 5 present matched accuracy and at-least-one accuracy for specialist referral. Several open-source SLMs showed strong zero-shot performance. Qwen2.5-7B-Instruct achieved the highest zero-shot matched accuracy among the SLMs at 68.19%, followed by Mistral-7B-Instruct at 63.90% and Llama-3.1-8B-Instruct at 63.76%. These results indicate that larger instruction-tuned SLMs could identify relevant specialties even without task-specific fine-tuning.

**Table 4:** Matched accuracy for referral performance comparison across zero-shot prompting, prefix tuning, LoRA, and full fine-tuning. Results for trainable methods are reported as mean ± standard deviation over five seeds.

| Model | Zero-shot | Prefix tuning | LoRA | Full fine-tuning |
|---|---|---|---|---|
| Llama-3.2-1B-Instruct | 42.41 | 56.96 ± 1.62 | 66.60 ± 2.82 | 69.68 ± 2.46 |
| Qwen2.5-1.5B-Instruct | 49.80 | 51.41 ± 3.93 | 60.93 ± 2.00 | 68.97 ± 1.09 |
| Qwen2.5-3B-Instruct | 50.44 | 64.99 ± 1.99 | 72.29 ± 2.39 | 72.77 ± 1.75 |
| Llama-3.2-3B-Instruct | 53.57 | 61.96 ± 1.97 | 71.96 ± 3.00 | 72.80 ± 2.56 |
| Qwen3-4B-Instruct | 60.60 | 65.53 ± 0.97 | 73.86 ± 1.92 | **75.22 ± 2.08** |
| Mistral-7B-Instruct | 63.90 | **68.66 ± 1.47** | **75.24 ± 1.59** | 73.97 ± 1.50 |
| Qwen2.5-7B-Instruct | 68.19 | 64.50 ± 2.77 | 71.88 ± 1.49 | 74.29 ± 2.19 |
| Llama-3.1-8B-Instruct | 63.76 | 64.20 ± 2.68 | 74.25 ± 2.31 | 73.98 ± 2.51 |
| Claude Haiku 4.5 | 70.68 | / | / | / |
| Claude Sonnet 4.5 | **71.04** | / | / | / |

Under the matched accuracy metric, several adapted open-source models surpassed both commercial baselines. The strongest result was achieved by Mistral-7B-Instruct with LoRA at 75.24% ± 1.59%, slightly ahead of Qwen3-4B-Instruct with full fine-tuning at 75.22% ± 2.08%. Llama-3.1-8B-Instruct with LoRA also performed strongly at 74.25% ± 2.31%, and Qwen2.5-3B-Instruct with LoRA reached 72.29% ± 2.39%. All these results exceeded Claude Sonnet 4.5, which achieved 71.04%, as well as Claude Haiku 4.5, which achieved 70.68%. Across model families, LoRA and full fine-tuning produced closely matched results for the 7B and 8B models, while prefix tuning generally remained lower, indicating that lightweight soft-prompt adaptation was less effective for this multi-label referral task than parameter updating methods.

**Table 5:** At-least-one accuracy for referral performance comparison across zero-shot prompting, prefix tuning, LoRA, and full fine-tuning. Results for trainable methods are reported as mean ± standard deviation over five seeds.

| Model | Zero-shot | Prefix tuning | LoRA | Full fine-tuning |
|---|---|---|---|---|
| Llama-3.2-1B-Instruct | 49.16 | 57.21 ± 1.81 | 68.39 ± 2.85 | 72.53 ± 2.56 |
| Qwen2.5-1.5B-Instruct | 61.20 | 63.56 ± 3.34 | 71.57 ± 2.61 | 80.29 ± 1.45 |
| Qwen2.5-3B-Instruct | 63.37 | 77.84 ± 1.68 | 83.08 ± 2.75 | 80.67 ± 2.52 |
| Llama-3.2-3B-Instruct | 64.82 | 62.12 ± 1.74 | 82.41 ± 2.63 | 78.07 ± 3.88 |
| Qwen3-4B-Instruct | 73.98 | 75.29 ± 2.37 | 85.35 ± 2.07 | 84.34 ± 2.68 |
| Mistral-7B-Instruct | 76.39 | **80.43 ± 1.21** | **87.04 ± 1.59** | **86.02 ± 1.38** |
| Qwen2.5-7B-Instruct | 72.05 | 75.38 ± 4.07 | 80.29 ± 1.72 | 81.25 ± 2.30 |
| Llama-3.1-8B-Instruct | 74.94 | 65.62 ± 4.18 | 86.41 ± 2.44 | 82.27 ± 4.35 |
| Claude Haiku 4.5 | 84.60 | / | / | / |
| Claude Sonnet 4.5 | **85.30** | / | / | / |

A similar pattern was observed under the at-least-one metric. Mistral-7B-Instruct with LoRA achieved the best overall result at 87.04% ± 1.59%, exceeding Claude Sonnet 4.5, which achieved 85.30%, and Claude Haiku 4.5, which achieved 84.60%. Mistral-7B-Instruct with full fine-tuning also performed strongly at 86.02% ± 1.38%, and Llama-3.1-8B-Instruct with LoRA reached 86.41% ± 2.44%. Qwen3-4B-Instruct with full fine-tuning remained competitive at 84.34% ± 2.68%, while Qwen3-4B-Instruct with LoRA achieved 85.35% ± 2.07% and slightly surpassed the Haiku baseline. Compared with matched accuracy, at-least-one accuracy was consistently higher across nearly all models, indicating that adapted open-source SLMs often retrieved at least one clinically relevant specialty even when they did not fully match the target set.

**Diagnosis Prediction** Of the three tasks, diagnosis prediction posed the greatest challenge for open-source SLMs, as shown in Tables 6 and 7. Among the open-source SLMs, Mistral-7B-Instruct achieved the highest zero-shot matched accuracy at 61.16%, followed by Qwen2.5-7B-Instruct at 56.14% and Llama-3.1-8B-Instruct at 53.25%. However, all zero-shot SLMs remained below Claude Haiku 4.5 at 71.73% and Claude Sonnet 4.5 at 74.38%.

**Table 6:** Matched accuracy for diagnosis performance comparison across zero-shot prompting, prefix tuning, LoRA, and full fine-tuning. Results for trainable methods are reported as mean ± standard deviation over five seeds.

| Model | Zero-shot | Prefix tuning | LoRA | Full fine-tuning |
|---|---|---|---|---|
| Llama-3.2-1B-Instruct | 38.31 | 49.42 ± 4.52 | 52.96 ± 2.84 | 50.10 ± 2.00 |
| Qwen2.5-1.5B-Instruct | 34.51 | 55.84 ± 2.85 | 55.88 ± 2.33 | 57.07 ± 2.28 |
| Qwen2.5-3B-Instruct | 45.12 | 57.90 ± 1.49 | 59.02 ± 2.59 | 57.51 ± 2.85 |
| Llama-3.2-3B-Instruct | 46.35 | 56.20 ± 1.18 | 59.96 ± 1.90 | 60.18 ± 5.69 |
| Qwen3-4B-Instruct | 49.06 | 61.12 ± 1.16 | 60.33 ± 1.88 | 66.22 ± 2.43 |
| Mistral-7B-Instruct | 61.16 | **68.84 ± 3.41** | **67.13 ± 2.14** | **67.55 ± 1.91** |
| Qwen2.5-7B-Instruct | 56.14 | 61.63 ± 2.94 | 61.65 ± 1.40 | 64.27 ± 2.31 |
| Llama-3.1-8B-Instruct | 53.25 | 59.48 ± 2.63 | 61.56 ± 2.00 | 63.22 ± 1.83 |
| Claude Haiku 4.5 | 71.73 | / | / | / |
| Claude Sonnet 4.5 | **74.38** | / | / | / |

We then examined whether adaptation improved matched diagnosis accuracy. Adaptation improved matched accuracy for most models. Mistral-7B-Instruct achieved the highest adapted SLM result with prefix tuning at 68.84% ± 3.41%, followed by full fine-tuning at 67.55% ± 1.91% and LoRA at 67.13% ± 2.14%. Qwen3-4B-Instruct with full fine-tuning also performed competitively at 66.22% ± 2.43%. Despite these gains, all adapted open-source models remained below Claude Sonnet 4.5, which reached 74.38%. This indicates that matched diagnosis accuracy represents the largest remaining gap between locally deployable open-source SLMs and strong proprietary commercial models in our benchmark.

**Table 7:** At-least-one accuracy for diagnosis performance comparison across zero-shot prompting, prefix tuning, LoRA, and full fine-tuning. Results for trainable methods are reported as mean ± standard deviation over five seeds.

| Model | Zero-shot | Prefix tuning | LoRA | Full fine-tuning |
|---|---|---|---|---|
| Llama-3.2-1B-Instruct | 38.8 | 50.07 ± 4.26 | 54.53 ± 2.62 | 50.12 ± 1.88 |
| Qwen2.5-1.5B-Instruct | 36.1 | 61.49 ± 3.23 | 63.65 ± 2.55 | 63.65 ± 2.34 |
| Qwen2.5-3B-Instruct | 48.19 | 62.45 ± 1.37 | 64.84 ± 2.59 | 60.58 ± 3.62 |
| Llama-3.2-3B-Instruct | 50.14 | 59.45 ± 1.34 | 62.16 ± 2.11 | 62.78 ± 6.39 |
| Qwen3-4B-Instruct | 52.51 | 62.93 ± 1.86 | 63.07 ± 2.37 | 69.98 ± 2.49 |
| Mistral-7B-Instruct | 63.61 | **73.96 ± 2.35** | **73.75 ± 0.79** | **73.96 ± 1.85** |
| Qwen2.5-7B-Instruct | 57.83 | 66.67 ± 3.51 | 67.00 ± 2.00 | 71.13 ± 2.77 |
| Llama-3.1-8B-Instruct | 58.07 | 61.97 ± 2.20 | 64.65 ± 1.70 | 67.43 ± 3.64 |
| Claude Haiku 4.5 | 82.34 | / | / | / |
| Claude Sonnet 4.5 | **84.10** | / | / | / |

We also evaluated at-least-one accuracy. Under this metric, Mistral-7B-Instruct again achieved the best open-source performance and showed highly consistent results across prefix tuning, LoRA, and full fine-tuning, reaching 73.96% ± 2.35%, 73.75% ± 0.79%, and 73.96% ± 1.85%, respectively. Qwen3-4B-Instruct with full fine-tuning reached 69.98% ± 2.49%, while Qwen2.5-7B-Instruct with full fine-tuning achieved 71.13% ± 2.77%, both representing substantial gains over their zero-shot baselines. However, even the strongest open-source result remained well below Claude Sonnet 4.5, which achieved 84.10%. Compared with triage and specialist referral, diagnosis therefore remained the most difficult task to close with local SLM adaptation. This gap may reflect the wider range of possible diagnoses and clinical terms, as well as differences introduced by the LLM-based evaluation method.

**Discussion**

This study presents a systematic comparison of prompting and fine-tuning strategies for eight open-source SLMs across three ED clinical decision-support tasks. Task-specific adaptation improved model performance across triage, specialist referral, and diagnosis prediction. In triage prediction, LoRA adapted Qwen3-4B-Instruct achieved the highest exact-match accuracy at 65.98%, exceeding Claude Sonnet 4.5, which reached 62.68%. Several adapted SLMs achieved higher exact-match accuracy than the commercial zero-shot baselines for triage and referral, although Claude Sonnet 4.5 retained the highest triage range accuracy. Diagnosis prediction remained more difficult, and all adapted SLMs performed below Claude Sonnet 4.5 on this task. These findings suggest that adapted open-source SLMs may be suitable for some ED decision-support tasks, but their performance depends on the task and evaluation metric.

Specialist referral showed the strongest results for the adapted SLMs, with several models achieving higher observed accuracy than both commercial baselines. In contrast, diagnosis prediction remained the most difficult task. This may be related to the wider range of possible diagnoses, variation in clinical terminology, and the use of an LLM judge for evaluation. Across the three tasks, LoRA produced consistently strong results for several model families, while prefix tuning was less consistent. LoRA also updates fewer parameters than full fine-tuning, although this study did not directly compare training time or memory use. Taken together, these findings support LoRA as the most practical default strategy, offering strong performance with relatively low computational cost.

The confusion matrix analysis also showed differences that were not captured by overall accuracy. Mistral-7B-Instruct with LoRA identified 8 of the 33 ESI-1 cases, whereas Claude Sonnet 4.5 identified none. However, both models had low sensitivity for ESI-1 cases, and the small sample size limits the interpretation of this result. Several other limitations should also be considered. The ESI distribution was imbalanced, the referral reference labels were generated by Claude Sonnet 4.5, and diagnosis evaluation relied on a Llama-3.1-70B judge. In addition, this study used data from a single clinical database and did not include external or prospective validation. Future studies should use labels reviewed by clinicians, evaluate performance across patient subgroups and healthcare settings, and examine how these models affect clinical decisions in practice.

## Conclusion

We conducted a systematic benchmark of adaptation strategies for eight open-source SLMs on three ED decision-support tasks. Task-specific adaptation improved performance across the evaluated models, and several adapted SLMs achieved higher observed accuracy than the commercial zero-shot baselines for triage exact match and specialist referral. However, Claude Sonnet 4.5 retained the highest triage range accuracy, and commercial models performed better on diagnosis prediction. LoRA produced consistently strong results across several models and tasks. Confusion matrix analysis showed that Mistral-7B-Instruct with LoRA identified some ESI-1 cases missed by Claude Sonnet 4.5, although both models had low sensitivity for this group. These findings demonstrate the feasibility of adapting open-source SLMs for local ED decision support and highlight their potential value for hospitals in protecting patient privacy and keeping clinical data within hospital systems.

## Acknowledgements

Q. Zhang and Y. Guo were partially supported by a seed grant from the UT San Antonio Office of Research and Innovation and NSF Grants CNS-2106761, CMMI-2222670, and CNS-2318683. Y. Gong was partially supported by NSF Grant CNS-2611068.